%% file: main.tex
\documentclass[pmlr,twocolumn,10pt]{jmlr} 

\mlhtrack{proceedings}

\newif\iffinal
\finaltrue  

\iffinal
    \ifmlhneedspmlr
      \jmlrvolume{346}
      \jmlryear{2026}
    \fi
    \ifmlhfindings \jmlrproceedings{}{ML4H 2026 - Findings Track}\fi
    \ifmlhdemo     \jmlrproceedings{}{ML4H 2026 - Demo Track}\fi
    \jmlrworkshop{Machine Learning for Health (ML4H) 2026}
\else
    \jmlrproceedings{}{Submitted to ML4H 2026: \mlhtrackname}
    \jmlrworkshop{Machine Learning for Health (ML4H) 2026}
\fi

\usepackage[switch]{lineno}
\usepackage[utf8]{inputenc} 
\usepackage[T1]{fontenc}    
\usepackage{hyperref}       
\usepackage{url}            
\usepackage{booktabs}       
\usepackage{amsfonts}       
\usepackage{nicefrac}       
\usepackage{microtype}      
\usepackage{xcolor}         
\usepackage{graphicx}
\usepackage{amsmath}
\usepackage{amssymb}
\usepackage{makecell}
\usepackage{booktabs}
\usepackage{multirow}
\usepackage{wrapfig}
\usepackage{float}
\usepackage{placeins}
\usepackage{fvextra}

\usepackage{tikz}
\usepackage{pgfplots} 
\usepackage{booktabs} 
\usetikzlibrary{shapes.geometric, arrows.meta, positioning, calc, matrix}
\pgfplotsset{compat=1.18}

\title{Evaluating Pretrained Language Semantics for Structured EHR Prediction Tasks}

\author{%
    \Name{Ashvin Gupta} \Email{ashvin.gupta19@imperial.ac.uk}\\
    \Name{Brendan Delaney} \Email{brendan.delaney@imperial.ac.uk}\\
    \Name{Alessandra Russo} \Email{a.russo@imperial.ac.uk}\\
    \addr Imperial College London, United Kingdom
}
\begin{document}

\maketitle

\input{text/abstract}

\input{text/availability}
\input{text/introduction}
\input{text/methods}
\input{text/experiments}
\input{text/results}

\input{text/discussion}

\bibliography{references}

\input{text/appendix}

\end{document}

%% file: text/abstract.tex
\begin{abstract}

Structured Electronic Health Record (EHR) models typically represent clinical concepts as dataset-specific tokens, discarding their associated natural-language semantics. Consequently, relationships between concepts must largely be learned from EHR co-occurrence without directly exploiting pretrained linguistic knowledge.
We investigate the contribution of pretrained language semantics to longitudinal EHR prediction through tEHR, a framework that converts clinical events into natural-language descriptions and adapts pretrained autoregressive language models through EHR-specific continued pretraining and task-specific discriminative fine-tuning.
We evaluate tEHR across four MIMIC-IV prediction tasks, where Mistral-7B achieves the strongest overall performance against structured-data, code-based, and LLM-based baselines. Controlled ablations demonstrate the importance of lexical grounding, numerical values, and both adaptation stages. A capacity-controlled experiment further shows that replacing clinical descriptions with atomic code tokens substantially reduces performance despite retaining the same LLM backbone.
Alternative approaches incorporating pretrained semantic embeddings offer lower computational costs but do not match full textualisation. These results highlight the importance of both semantic representation and downstream adaptation when applying pretrained language models to structured EHR prediction.


\end{abstract} 


%% file: text/availability.tex
\ifmlhneedsstatements
\paragraph*{Data and Code Availability}
We use the publicly available MIMIC-IV dataset \citep{Johnson2024MIMIC-IV}
, processed into the Medical Event Data
Standard (MEDS) format \citep{McDermott2025MIMIC-IVFormat} with downstream cohorts assembled using the ACES framework \citep{Xu2025ACES:Datasets}.
Code can be found at \url{https://github.com/Ashvin-Gupta/tEHR}.

\paragraph*{Institutional Review Board (IRB)}
This research uses the publicly available MIMIC-IV dataset, which has been previously approved for research use and does not require additional IRB approval. 

%% file: text/introduction.tex
\section{Introduction}
Electronic Health Records (EHRs) provide longitudinal records of diagnoses, medications, procedures, laboratory measurements, and other clinical events. Transformer-based architectures have achieved strong performance across tasks including early pancreatic cancer detection \citep{Placido2023ATrajectories, Steinberg2023MOTOR:Records} and ICU outcome prediction \citep{Harutyunyan2019MultitaskData}, with recent work extending to generative sequence models \citep{Kraljevic2023ForesightEHRs,Wornow2023EHRSHOT:Models, Renc2024ZeroTransformer, Shmatko2025LearningTransformers, Rajamohan2025FoundationScale, Waxler2025GenerativeScale}, state-space models \citep{Fallahpour2024EHRMamba:Records}, and prompt-based prediction \citep{Chandak2026EveryQuery:Records}.

Despite their architectural differences, these approaches largely rely on each clinical event being expressed as unique tokens. Consequently, models must learn representations and relationships between concepts from their EHR co-occurrence, and cannot exploit any prior semantic knowledge. These clinical concepts are recorded using a range of coding systems, including SNOMED-CT \citep{Stearns2001SNOMEDStatus.}, ICD \citep{WorldHealthOrganizationInternationalICD}, and Read Codes \citep{Chisholm1990TheClassification.}, with the choice of vocabulary varying across healthcare systems. Even within a single coding system, the subset of codes used can differ between sites and evolve over time as coding practices and terminologies change. Consequently, the same underlying clinical concept may be represented by multiple codes across datasets, while the overall vocabulary can contain thousands of possible concepts: Non-Type 1 Diabetes Mellitus is represented by $768$ out of $>360,000$ active concepts in SNOMED-CT \citep{NHSEnglandPrimaryCareDomain2025Non-TypeDMNONTYPE1_COD}. This heterogeneity often requires ontology mapping, code collapsing, or vocabulary refinement before modelling \citep{Wang2024Transformer-basedData, Makarov2025ARecords}, creating a dataset-specific vocabulary and model. Transferring a model to a new dataset therefore requires mapping clinical codes to the expected vocabulary of the model, which can lead to information loss. Furthermore, models take different approaches to representing the time intervals between events or continuous values of recordings \citep{Guo2026TokenizationModels, Attrach2025RethinkingMore, Lee2026RepresentationModels}. These preprocessing requirements make structured-EHR modelling difficult to reuse consistently across datasets and healthcare systems.

\begin{figure*}[h!]
    \centering
    \includegraphics[width=1\linewidth]{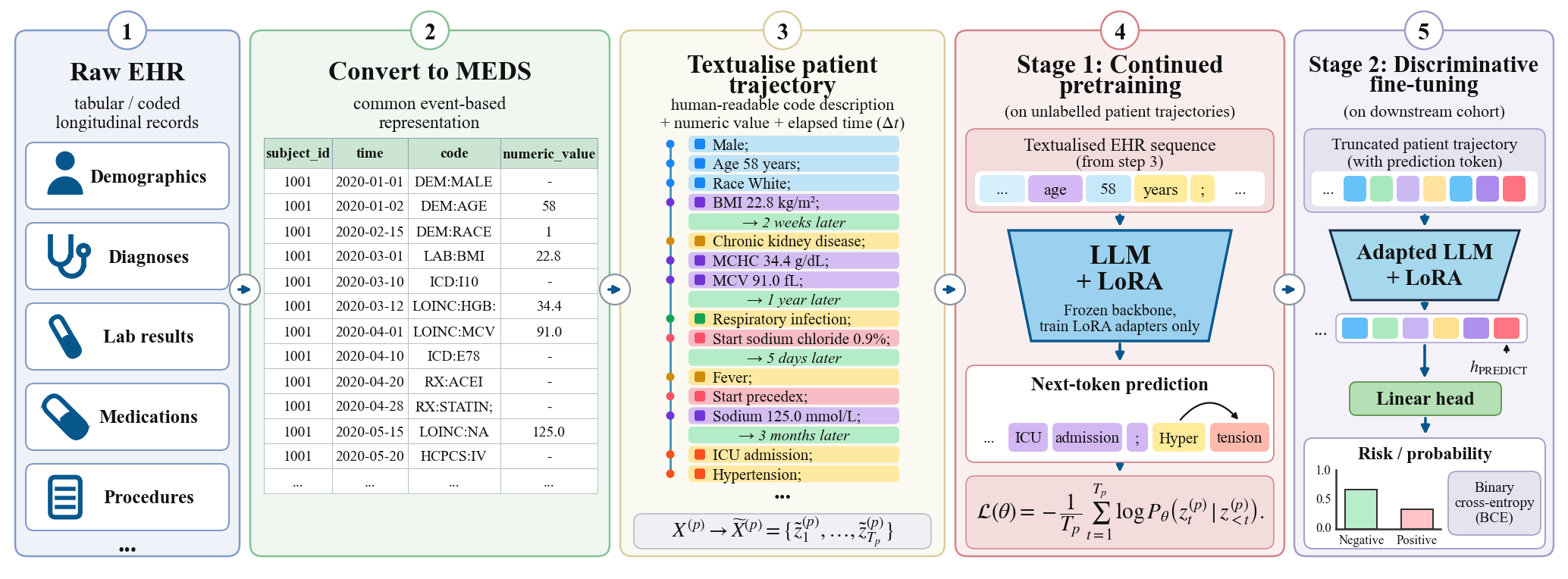}
    \caption{Structured EHRs are converted into the MEDS event-based representation \citep{BertArnrich2024MedicalHealth} and textualised using natural-language descriptions of clinical events, numerical values, and elapsed time between events. A pretrained large language model (LLM) is adapted to unlabelled patient trajectories using next-token prediction, with low-rank adaptation (LoRA) while keeping the backbone frozen. For downstream prediction tasks, patient trajectories are truncated at the prediction time and appended with a $[\mathrm{PREDICT}]$ token. The adapted LLM produces a contextual representation of this token, which is passed to a linear classification head to estimate the probability of the target outcome using binary cross-entropy loss.}
    \label{fig:overall approach}
\end{figure*}

Replacing symbolic clinical codes with their natural-language descriptions provides a direct way to exploit the semantic knowledge encoded in pretrained language models and reduce the need for vocabulary-specific preprocessing. Recent work suggests that these representations can improve structured EHR prediction, but it remains unclear how pretrained language semantics are beneficial for longitudinal modelling and how the language model should be incorporated into the prediction pipeline. Prompt-based approaches present patient histories directly to an LLM, either without parameter updates \citep{Zhu2024PromptingData} or with supervised fine-tuning \citep{Shoham2024CPLLM:Models}, treating prediction primarily as a language-generation problem. Other work uses pretrained language models as representation encoders. Event-level embedding approaches encode each clinical event and pass these through a separate longitudinal aggregator, typically a Transformer, for the final prediction \citep{Hur2024UniHPFKnowledge,Park2025EnhancingEmbeddings}. Patient-level embedding approaches encode the full patient trajectory into a single representation that is subsequently used by a lightweight classifier, using either clinically pretrained encoders \citep{Munoz-Farre2022SEHR-CE:Expansion} or general-purpose embedding-specific models \citep{Hegselmann2026LargeEncoders}. These studies establish that natural-language representations can provide useful inductive bias for clinical prediction tasks, but leave unresolved how much performance comes from the pretrained semantics themselves.


To investigate these questions, we develop tEHR, a two-stage framework for adapting pretrained autoregressive LLMs to longitudinal EHR prediction. tEHR converts structured clinical events into natural-language descriptions, applies continued pretraining on unlabelled EHR trajectories, and performs task-specific discriminative fine-tuning using the LLM as a longitudinal patient encoder.

We use this framework to investigate whether pretrained lexical knowledge contributes to predictive performance beyond model capacity alone, which components of the clinical representation are important, and whether alternative approaches to incorporating pretrained semantic information can achieve comparable performance at lower computational cost.
 
Our contributions are:




\begin{enumerate}
    \item We present tEHR, a framework for adapting
    pretrained autoregressive LLMs to structured EHR
    prediction through continued pretraining on
    longitudinal patient trajectories followed by
    task-specific discriminative fine-tuning.

    \item Through controlled ablations, we investigate
    the contributions of pretrained lexical grounding,
    numerical and temporal information, model capacity,
    and the two stages of adaptation to predictive
    performance.

    \item We evaluate alternative approaches to
    incorporating pretrained language semantics into
    longitudinal EHR models, comparing predictive
    performance and computational cost across four
    MIMIC-IV prediction tasks.
\end{enumerate}

%% file: text/methods.tex
\section{Methodology}

\subsection{Patient Trajectory Representation}
A patient trajectory, $X^{(p)}$, is represented as a time-ordered sequence of clinical events,
$X^{(p)}=(e^{(p)}_1, e^{(p)}_2, \ldots, e^{(p)}_{L_p})$,
where $L_p$ denotes the number of clinical events for patient $p$. Each event is represented as
$e^{(p)}_i = (c^{(p)}_i, v^{(p)}_i, t^{(p)}_i)$,
where $c^{(p)}_i \in \mathcal{V}$ is a discrete medical code, $v^{(p)}_i \in \mathbb{R}$ is an optional continuous numerical value, and $t^{(p)}_i$ is the event timestamp. Static patient attributes, such as sex and year of birth, are retained at the beginning of the patient sequence.

\subsection{Clinical Narrative Textualisation}
The EHR dataset is first transformed into the Medical Event Data Standard (MEDS) \citep{BertArnrich2024MedicalHealth}, which provides a common event-based representation across coding systems. We then apply a deterministic textualisation function that converts each structured patient trajectory into a sequence of natural-language event descriptions:

\begin{equation}
X^{(p)}
\longrightarrow
\tilde{X}^{(p)}
=
\left(
\tilde{x}^{(p)}_1,\ldots,\tilde{x}^{(p)}_{L_p}
\right),
\end{equation}

where $\tilde{x}^{(p)}_i$ denotes the textual representation of event $e^{(p)}_i$.

For each event $e^{(p)}_i$, the symbolic concept $c^{(p)}_i$ is replaced by its human-readable description using official coding-system translation metadata. Numerical measurements are retained in the resulting text and rounded to one decimal place. Elapsed time between consecutive events is also included in the text stream, $\Delta t^{(p)}_i = t^{(p)}_i - t^{(p)}_{i-1}$,
using the nearest clinically interpretable unit, such as minutes, hours, days, weeks, months, or years. 
Conceptually, the representation remains a time-ordered event sequence, but with all events expressed through their natural-language description together with any associated numerical value or time delta, rather than as a single code token.
Unlike code-based approaches, it does not require a dataset-specific clinical vocabulary, frequency-based code pruning, or specialised temporal and numerical embedding modules.

\subsection{Two-Stage Training Framework}
The tEHR framework adapts a causal LLM in two stages. First, the LLM is fine-tuned on the unlabelled patient trajectories through continued pretraining. The second involves discriminative fine-tuning with a prediction head on top of a specific hidden representation for the downstream task of interest. Both stages use Low-Rank Adaptation (LoRA) \citep{Hu2021LoRA:Models, Schulman2025LoRARegret} applied to the self-attention and linear projection layers.
For a pre-trained weight matrix $W_0 \in \mathbb{R}^{d \times k}$, LoRA represents the update $\Delta W$ as the product of two low-rank matrices $B \in \mathbb{R}^{d \times r}$ and $A \in \mathbb{R}^{r \times k}$, such that the modified forward pass is: $h = (W_0  + \Delta W) x  = (W_0 + \frac{\alpha}{r} BA)x$, where $\alpha$ is the scaling constant and $r$ is the rank. During both stages of training, $W_0$ remains frozen while only $A$ and $B$ are updated, substantially reducing the number of trainable parameters while retaining the pretrained representations of the backbone model.

\paragraph{Stage 1: Domain-adaptive continued pretraining.}

The model is trained on the textualised training set patient trajectories using the standard next-token prediction loss, updating the LoRA parameters $\theta$. The textual sequence $\tilde{X}^{(p)}$ is tokenised using the LLM's built-in tokeniser, producing
$Z^{(p)}=z^{(p)}_1,\ldots,z^{(p)}_{T_p}$, 
where $T_p$ denotes the number of LLM tokens in the tokenised trajectory for patient $p$. Note that $T_p$ differs from $L_p$, since each clinical event description may be represented by multiple language-model tokens.

\begin{equation}
\mathcal{L}_{\mathrm{CLM}}(\theta)
=
-\frac{1}{T_p}
\sum_{t=1}^{T_p}
\log P_{\theta}
\left(
z_t^{(p)} \mid z_{<t}^{(p)}
\right).
\end{equation}

This stage is intended to align the pretrained model with the distribution of longitudinal EHR trajectories without downstream outcome labels. 

\paragraph{Stage 2: Discriminative fine-tuning.}

For each downstream task, the adapted model receives the patient narrative truncated at the corresponding prediction time, appended with a special $\mathrm{[PREDICT]}$ token. The hidden representation of this token, $h^{(p)}_\mathrm{{PREDICT}} \in \mathbb{R}^d$, is input to a linear classifier layer to produce the probability $\hat{y}_p =\sigma (w^\top h^{(p)}_{\mathrm{PREDICT}} + b)$ with the model further fine-tuned under the Binary Cross-Entropy objective:

\begin{equation}
\mathcal{L}_{\mathrm{BCE}}=
-\frac{1}{N}
\left[
\sum_{p=1}^{N}
y_p \log \hat{y}_p
+
(1-y_p)\log(1-\hat{y}_p)
\right]
\end{equation}

for $N$ patients. This stage updates the task-specific classification head and the LoRA adapters inherited from Stage~1, while the original backbone weights remain frozen. We use a linear head so that downstream performance primarily reflects the quality of the learned patient representation. The hidden state of the $\mathrm{[PREDICT]}$ token is used as it outperformed alternative pooling strategies such as mean pooling in preliminary experiments. Both adaptation stages are implemented with Unsloth \citep{DanielHan2023Unsloth} and trained with early stopping on validation loss using a fixed 80:10:10 patient-level split and three random seeds. 

%% file: text/experiments.tex
\section{Experiments}
\subsection{Dataset and Prediction Tasks}


\paragraph{MIMIC-IV:}
We use MIMIC-IV version 3.1 \citep{Johnson2024MIMIC-IV}, a large critical-care dataset containing events recorded during hospital and intensive-care admissions. We assess four binary prediction tasks: ICU mortality, ICU readmission, hospital mortality, and 30-day hospital readmission. MIMIC-IV contains clinical concepts represented primarily using ICD and NDC coding systems. In our processed dataset, this corresponds to roughly $15k$ unique medical codes. It provides a high-acuity evaluation setting characterised by dense event sequences and comparatively short prediction horizons.


\subsection{Baselines}
We group our baselines into structured code-based models and semantic representation-based prediction approaches. 

\paragraph{Structured Code-Based Models:}
\textbf{XGBoost} is trained on bag-of-codes frequency features. \textbf{Transformer Encoder} following \citet{Placido2023ATrajectories} is trained on the corresponding symbolic event sequences with an analogous pretraining and supervised fine-tuning procedure. \textbf{CLMBR-T} \citep{Wornow2023EHRSHOT:Models}, is a 141M-parameter structured-EHR Transformer pretrained on 2.57 million patients. Applying CLMBR-T to our datasets requires mapping the dataset codes to the model's vocabulary.
Using the public MIMIC-IV-to-OMOP mapping pipeline \citep{OHDSIMIMICConversion}, 79.4\% of event occurrences are mapped. 
We evaluate both a frozen CLMBR-T representation with a trainable linear-classifier head and a task-specific fine-tuned variant.

\paragraph{Semantic Representation Models:}
\textbf{BioClinical ModernBERT} is fine-tuned on the same textualised patient trajectories as tEHR \citep{Sounack2025BioClinicalNLP}.
We evaluate \textbf{zero-shot LLM} prompting of an LLM following \cite{Zhu2024PromptingData} and \textbf{prompt-based supervised fine-tuning} of an LLM following \citep{Shoham2024CPLLM:Models} 
.
We investigate three alternative approaches to incorporating pretrained language semantics into longitudinal EHR prediction. First, \textbf{LLM-initialised Transformer}, motivated by \citet{Park2025EnhancingEmbeddings}, uses a code-based Transformer whose clinical-code embeddings are initialised from pretrained LLM representations of their corresponding natural-language descriptions.
Second, \textbf{Event LLM embeddings}, using a hierarchical architecture similar to \citet{Hur2024UniHPFKnowledge}, represents individual clinical events with frozen pretrained LLM embeddings, which are passed to a trainable Transformer for longitudinal prediction. Third, \textbf{LLM projection} projects pretrained event-level LLM embeddings into the input embedding space of a longitudinal LLM. The latter two approaches use task-specific supervised training without the autoregressive next-token continued pretraining stage employed by tEHR. We report AUROC, AUPRC and computational cost in the main text with full implementation details provided in Appendix \ref{Baselins_and_Ablations}. 

\begin{table*}[t]
\centering
\caption{
AUROC and AUPRC across four MIMIC-IV prediction tasks,
averaged over three seeds. Best results are shown in bold.
Standard deviations are reported in the appendix.
}
\label{tab:mimic_auroc}

\begingroup
\footnotesize
\setlength{\tabcolsep}{5pt}

\begin{tabular}{@{}lcccccccc@{}}
\toprule
& \multicolumn{2}{c}{\textbf{ICU Mortality}}
& \multicolumn{2}{c}{\textbf{ICU Readm.}}
& \multicolumn{2}{c}{\textbf{Hospital Mortality}}
& \multicolumn{2}{c}{\textbf{Hospital Readm.}} \\
\cmidrule(lr){2-3}
\cmidrule(lr){4-5}
\cmidrule(lr){6-7}
\cmidrule(l){8-9}

\textbf{Model}
& AUROC & AUPRC
& AUROC & AUPRC
& AUROC & AUPRC
& AUROC & AUPRC \\
\midrule

XGBoost
& 0.817 & 0.306
& 0.713 & 0.333
& 0.868 & 0.235
& 0.742 & 0.483 \\

Transformer Encoder
& 0.866 & 0.370
& 0.773 & 0.456
& 0.911 & 0.281
& 0.765 & 0.531 \\

CLMBR-T (frozen)
& 0.793 & 0.235
& 0.724 & 0.340
& 0.881 & 0.204
& 0.708 & 0.409 \\

CLMBR-T (fine-tuned)
& 0.861 & 0.406
& 0.762 & 0.428
& 0.915 & 0.328
& 0.762 & 0.515 \\

BioClinical ModernBERT
& 0.824 & 0.300
& 0.739 & 0.359
& 0.860 & 0.197
& 0.766 & 0.532 \\

Zero-shot LLM (Mistral-7B)
& 0.559 & 0.109
& 0.539 & 0.147
& 0.627 & 0.048
& 0.528 & 0.232 \\

Prompt SFT (Mistral-7B)
& 0.811 & 0.221
& 0.668 & 0.219
& 0.786 & 0.073
& 0.724 & 0.437 \\

Mistral-7B (atomic codes)
& 0.772 & 0.221
& 0.703 & 0.323
& 0.871 & 0.226
& 0.766 & 0.511 \\

LLM-initialised Transformer
& 0.843 & 0.351
& 0.778 & 0.467
& 0.897 & 0.275
& 0.766 & 0.532 \\

Event LLM embeddings
& 0.739 & 0.194
& 0.547 & 0.166
& 0.876 & 0.215
& 0.739 & 0.480 \\

LLM projection
& 0.737 & 0.210
& 0.651 & 0.273
& 0.817 & 0.156
& 0.737 & 0.483 \\

\midrule

tEHR-Qwen3-0.6B
& 0.822 & 0.289
& 0.738 & 0.414
& 0.903 & 0.273
& 0.759 & 0.522 \\

tEHR-Qwen3-8B
& 0.870 & 0.391
& 0.780 & 0.480
& 0.927 & 0.324
& 0.764 & 0.527 \\

\textbf{tEHR-Mistral-7B}
& \textbf{0.887} & \textbf{0.424}
& \textbf{0.794} & \textbf{0.498}
& \textbf{0.932} & \textbf{0.333}
& \textbf{0.775} & \textbf{0.556} \\

\bottomrule
\end{tabular}

\endgroup
\end{table*}

\subsection{Ablations}
\label{ablation_method}

We conduct controlled ablations across the four MIMIC-IV prediction tasks to investigate three questions: (1) which stages of the tEHR training procedure are necessary, (2) which components of the textualised EHR representation contribute predictive information, and (3) how sensitive performance is to backbone choice and model capacity.


\paragraph{Two-stage training framework.}
We evaluate three training variants to disentangle the contributions of domain-adaptive continued pretraining and task-specific adaptation. (1) we remove Stage~1 and initialise directly from the base LLM, applying the same supervised LoRA adaptation and linear classification head used in the full model. (2) we retain Stage~1 but freeze the adapted backbone during downstream training, optimising only the linear classification head. (3) we freeze the initial unadapted base LLM and train only the classification head. Together, these variants separate the contribution of unlabelled EHR adaptation, task-specific backbone adaptation, and generic pretrained representations.

\paragraph{Clinical information representation.}
We next test which components of the textualised EHR representation contribute to predictive performance using controlled perturbations. In the \emph{hashed-concept} condition, each clinical concept is replaced by a deterministic, unique, and semantically meaningless identifier, preserving concept identity and recurrence structure while removing lexical meaning. In the \emph{scrambled-concept} condition, concept descriptions are deterministically permuted within event categories for the events that occurred for that patient, preserving plausible medical language and broad event type while breaking the correspondence between each concept and its true clinical meaning.
In the \emph{no-time} condition, explicit textualised elapsed-time intervals are removed while chronological event order is preserved. In the \emph{no-value} condition, numerical measurement values are removed while retaining the identity of the corresponding laboratory or vital-sign event. Across all conditions, demographics, hospital-transfer events and event ordering are preserved, and only the targeted component of the representation is modified.

\paragraph{Backbone and model capacity.}
We evaluate the full two-stage tEHR framework using Mistral-7B-v0.3, Qwen3-8B, and Qwen3-0.6B to assess robustness across LLM families and sensitivity to model capacity. To distinguish the contribution of pretrained lexical information from model capacity, we additionally construct a capacity-controlled variant using the Mistral-7B backbone, \textbf{Mistral-7B atomic codes}, replacing natural-language event descriptions with atomic clinical-code tokens. The vocabulary is extended with the clinical codes used by the structured baselines, whose embeddings are newly initialised. This variant follows the same two-stage training framework as tEHR, preserving backbone capacity and the two-stage adaptation procedure while replacing pretrained lexical representations with newly initialised code embeddings.

We use a maximum context length of 16,000 language-model tokens for tEHR and 2,048 clinical events for the code-based models where supported by their input representations and context limits. To ensure matched clinical histories, we apply a shared event-level truncation: for each patient, we retain the most recent events that fit within both limits, while preserving demographic information. The resulting identical event subset is provided to each model using its corresponding representation. BioClinical ModernBERT and CLMBR-T are the exceptions due to their $8,192$ and $496$ context legnth respectively. Token expansion and context-length characteristics are detailed in Appendix~\ref{context_length}.

%% file: text/results.tex
\section{Results}
\label{Results}

tEHR-Mistral achieved the highest AUROC and AUPRC across all four MIMIC-IV prediction tasks (Table~\ref{tab:mimic_auroc}). Improvements were particularly pronounced for ICU mortality, where tEHR achieved an AUPRC of 0.424 compared with 0.406 for fine-tuned CLMBR-T and 0.370 for the code-based Transformer. In contrast, zero-shot prompting performed substantially worse, with AUROC ranging from 0.528 to 0.627.

\begin{figure*}
    \centering
    \includegraphics[width=1\linewidth]{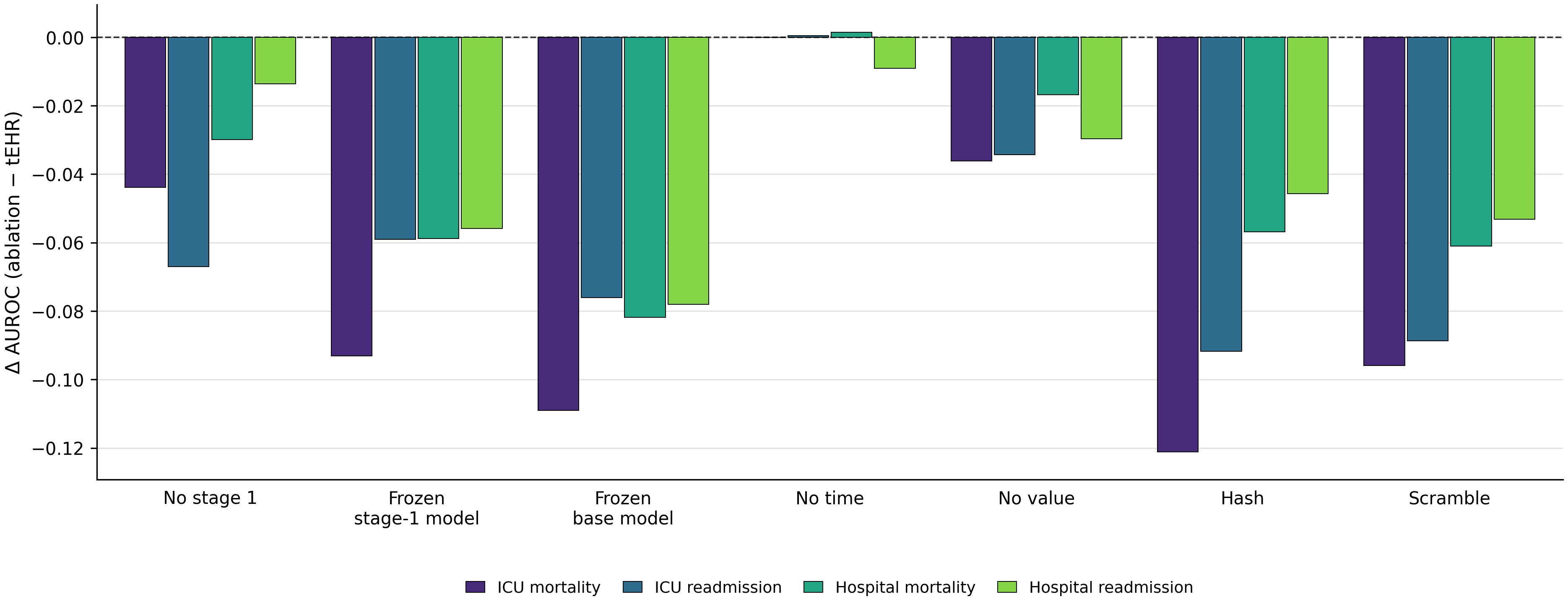}
    \caption{
    Change in AUROC relative to the full tEHR-Mistral-7B model across four downstream tasks on MIMIC-IV for the ablation experiments.}
    \label{fig:ablations}
    \vspace{-10pt}
\end{figure*}

The training ablations demonstrate the contribution of both adaptation stages (Figure~\ref{fig:ablations}). Removing EHR-specific continued pretraining reduced AUROC across all four tasks, while freezing the backbone during downstream training also reduced performance. Using frozen representations from the unadapted base LLM produced the weakest results among the training variants, demonstrating the importance of adapting the pretrained model to longitudinal clinical data and downstream prediction.

The clinical representation ablations further demonstrate the importance of lexical grounding. Replacing clinical descriptions with hashed identifiers or scrambling their correspondence to the underlying events substantially reduced performance across all tasks. Removing numerical values also consistently reduced AUROC, whereas removing explicit elapsed-time descriptions had little effect.

Performance varied across pretrained backbones, with Mistral-7B outperforming Qwen3-8B and Qwen3-0.6B. However, the capacity-controlled Mistral-7B experiment provides further evidence that model size alone does not explain tEHR's performance. Replacing natural-language descriptions with atomic clinical-code tokens reduced ICU mortality AUROC from 0.887 to 0.772, despite retaining the same backbone and two-stage training framework. Similar reductions were observed across the remaining tasks.

\begin{figure}
    \centering
    \includegraphics[width=1\linewidth]{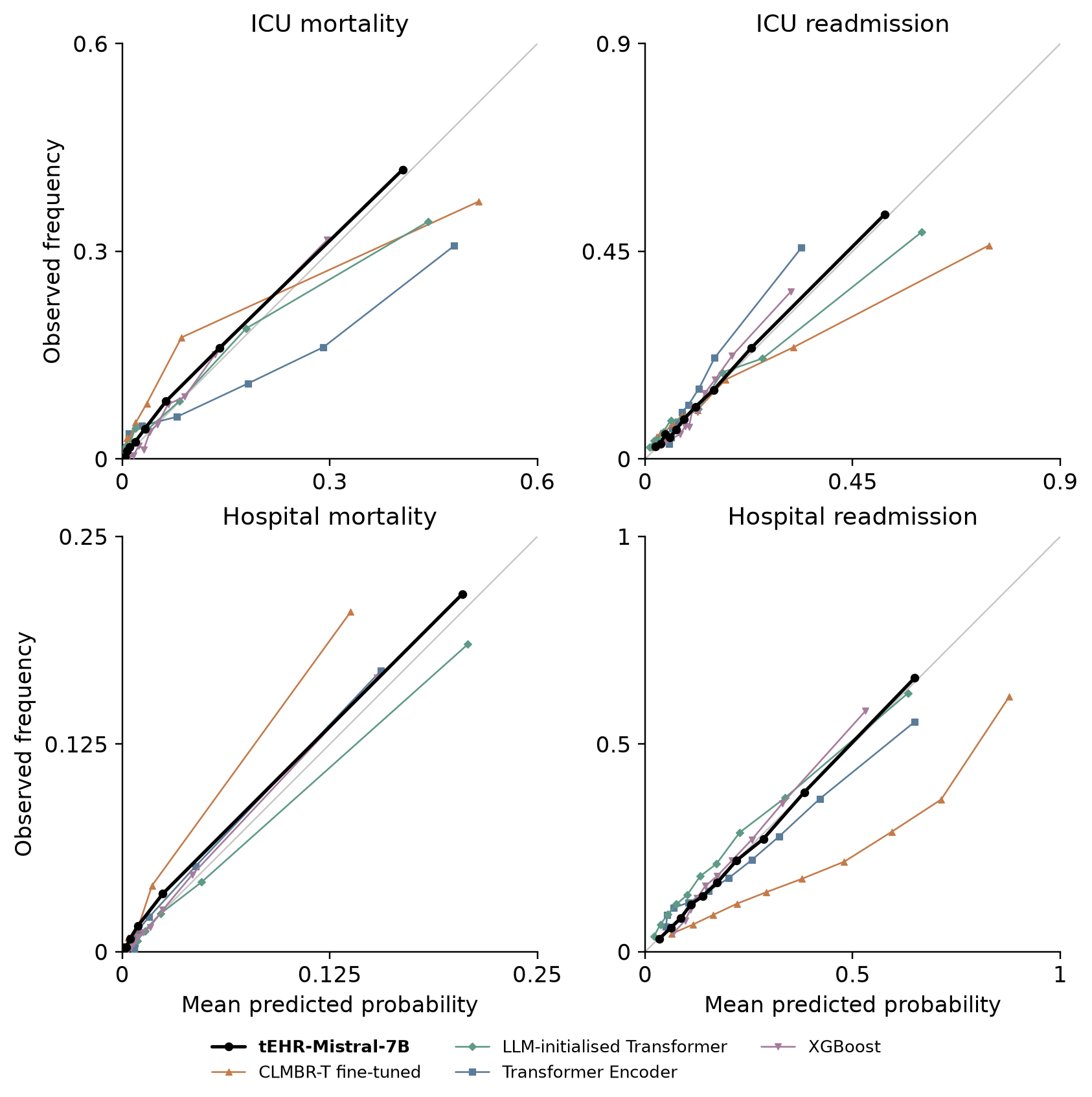}
    \caption{Calibration of predicted risks across four MIMIC-IV prediction tasks. Points show mean predicted probability and observed outcome frequency within 10 equal-count bins; connecting lines are included for visualisation only. The grey diagonal indicates perfect calibration. Axes use task-specific limits with equal ranges on both dimensions.}
    \label{figure:calibration}
\end{figure}

All alternative approaches to incorporating pretrained semantic representations performed
below full tEHR. The LLM-initialised Transformer was the strongest of these alternatives, achieving an ICU mortality AUROC of 0.843 compared with 0.887 for tEHR-Mistral. Event-level LLM embeddings and projected event representations yielded further reductions, suggesting that compact semantic representations did not recover the performance of full textualisation under the evaluated setups.

We additionally evaluated calibration across the four
MIMIC-IV tasks (Figure~\ref{figure:calibration}).
tEHR-Mistral achieved the lowest Brier score across
all tasks, although this metric reflects both
discrimination and calibration. Calibration was
particularly strong for ICU and hospital readmission,
with slopes of 1.01 and 1.03 and ECE values of
0.006 and 0.005, respectively. Other models achieved
better calibration on some metrics, particularly for
the mortality tasks (Appendix~\ref{app:calibration}).

\begin{table}[h]
\centering
\caption{
Computational cost on one NVIDIA H200. EHR PT denotes EHR-specific pretraining, and Task FT denotes supervised task training (ICU mortality), both in GPU-hours. Latency is measured per patient at batch size 1. External pretraining is excluded.
}
\label{tab:mimic_compute}

\begingroup
\scriptsize
\setlength{\tabcolsep}{3pt}

\begin{tabular}{@{}lrrr@{}}
\toprule
\textbf{Model}
& \shortstack{\textbf{EHR PT}\\\textbf{GPU-h}}
& \shortstack{\textbf{Task FT}\\\textbf{GPU-h}}
& \shortstack{\textbf{Latency}\\\textbf{ms}} \\
\midrule

XGBoost
& -- & -- & 0.21 \\
Transformer Encoder
& 96.11 & 15.65 & 8.8 \\
CLMBR-T (frozen)
& Ext. & 0.14 & 3.9 \\
CLMBR-T (fine-tuned)
& Ext. & 1.52 & 4.0 \\
BioClinical-ModernBERT
& Ext. & 3.09 & 487 \\
Zero-shot LLM
& -- & -- & 373 \\
Prompt SFT
& -- & 28.68 & 385 \\
Mistral-7B (atomic codes)
& 5.45 & 4.67 & 87.7 \\
LLM-initialised Transformer
& 96.10 & 15.65 & 8.75 \\
Event LLM embeddings
& -- & 16.03 & 8.86 \\
LLM projection
& -- & 4.73 & 77.7 \\

\midrule
tEHR-Qwen3-0.6B
& 19.66 & 6.62 & 68.0 \\
tEHR-Qwen3-8B
& 90.70 & 28.04 & 328 \\
\textbf{tEHR-Mistral-7B}
& 85.11 & 26.09 & 301 \\

\bottomrule
\end{tabular}
\endgroup
\end{table}

These performance differences were accompanied by computational trade-offs (Table~\ref{tab:mimic_compute}). tEHR-Mistral required 85.11 GPU-hours for EHR-specific continued pretraining and 26.09 GPU-hours for ICU mortality fine-tuning, compared with 96.11 and 15.65 GPU-hours for the code-based Transformer. However, single-patient inference was substantially slower (301 versus 8.8 ms), while the LLM-initialised Transformer achieved comparable latency to the code-based Transformer. Thus, more compact representations offered lower inference costs, but did not match tEHR's predictive performance.



%% file: text/discussion.tex
\section{Discussion}

We present tEHR, a framework for adapting pretrained autoregressive LLMs to EHR prediction tasks. tEHR can in principle be applied to different autoregressive LLMs and relies on clinical events being represented with their natural-language descriptions rather than as symbolic tokens, allowing the model to reuse semantic information learnt during large-scale pretraining. tEHR requires simple data preprocessing, avoids the construction of a model-specific vocabulary and can still be applied to large models through parameter-efficient fine-tuning with LoRA. 
However, the other LLM-based experiments show that simply introducing pretrained semantic embeddings does not recover the performance of full textualisation: 
zero-shot prompting performed poorly,
while alternative approaches incorporating
pretrained semantic embeddings also failed to
match full tEHR. This suggests that access to
pretrained language knowledge alone is insufficient,
and that how this knowledge is incorporated into
longitudinal modelling matters.

The ablations provide a possible explanation for the success of tEHR. Continued pretraining adapts the model to the distribution and sequential structure of the EHR data, while task-specific fine-tuning further shapes the representation for the downstream prediction task. 
The frozen base-model ablation assesses the predictive utility of generic pretrained representations without EHR-specific adaptation. The improvements obtained through continued pretraining and task-specific fine-tuning suggest that both stages contribute to learning effective longitudinal patient representations.
The lexical ablations further demonstrate the importance of correctly grounding clinical events in natural-language.
In \textit{hashed-concept}, the model is unable to identify the semantic meaning of the events occurring and performs worse. 
Similarly, the \textit{scrambled-concept} ablation indicates that it is not sufficient for the input to contain plausible clinical language but that the correspondence between a clinical event and its surrounding context matters.
Together, these perturbations suggest that the model benefits from the correspondence between clinical concepts and their pretrained lexical representations. This is further supported by the capacity-controlled Mistral-7B experiment, where replacing natural-language descriptions with atomic clinical-code tokens substantially reduced performance despite retaining the same backbone and two-stage training framework. Together, these findings suggest that pretrained lexical knowledge contributes to tEHR's performance beyond model capacity alone.

A possible explanation for the drop in performance when the clinical values are removed is that the model can no longer directly distinguish normal from abnormal measurement values. This is not the same case for time, consistent with previous work \citep{Attrach2025RethinkingMore} and suggesting that chronological ordering may therefore capture much of the required temporal information.
Performance also varied across pretrained backbones. Differences between Mistral-7B and Qwen3-8B, despite similar parameter counts, suggest that pretraining differences may influence downstream performance, although these experiments do not isolate their source. The weaker performance of Qwen3-0.6B further indicates sensitivity to model capacity.
These performance improvements come with a computational trade-off as tEHR was an expensive method for training and slower for inference, albeit with one patient prediction still comfortably under one second at 0.3s. More compact semantic representation strategies offered lower training costs, albeit with reduced predictive performance.

The CLMBR-T experiment illustrates a practical limitation of structured-EHR foundation models with fixed vocabularies. Although CLMBR-T performs strongly within its intended representation setting, applying it to our MIMIC-IV cohorts required mapping source concepts into the vocabulary expected by the model. Only 79.4\% of event occurrences were successfully mapped, which may partly explain the remaining performance gap. 
The mapping issue is of high concern and as such, the tEHR experiments were also conducted using a private UK primary care dataset for 12-month pancreatic cancer prediction. Primary-care is a substantially different setting from MIMIC-IV as is longitudinal data 
recorded across many practices and therefore results in a much larger and more heterogeneous clinical vocabulary; over $241k$ unique medical codes. A common preprocessing step is to map codes to higher-level clinical concepts, which is manually intensive and can lead to incorrect medical timelines if there are mistakes.
tEHR showed the strongest performance, showing that predictive models can be developed without extensive code simplification. A mapping of CLMBR-T's vocabulary to the private dataset vocabulary was not found meaning the model could not be run, highlighting the issue with many of the current code-based models. 
In contrast, tEHR operates on natural-language concept descriptions, avoiding explicit alignment to a pretrained symbolic vocabulary. 
An interesting future experiment would be to evaluate the cross-dataset transfer of a tEHR model which would complement recent work by \citet{Ding2026DiscreteModeling} on unifying vocabularies across datasets.

Several limitations remain. First, evaluation is restricted to four prediction tasks within MIMIC-IV and a private UK dataset which could not be shared. Further evaluation across independent healthcare systems, coding vocabularies, and prediction tasks is necessary to establish generalisability. Second, textualisation expands clinical events into multiple subword tokens, increasing computational cost and requiring truncation of longer patient histories.
Future work could investigate cross-dataset transfer
and the interpretability of learned representations.

\section{Acknowledgements}
This work is supported by UK Research and Innovation (UKRI AI Centre for Doctoral Training in Digital Healthcare grant number EP/Y030974/1), NIHR Imperial Biomedical Research Centre and the NIHR Invention for Innovation grant ``Artificial Intelligence to support cancer early diagnosis in general practice" (NIHR207533).

%% file: text/appendix.tex
\onecolumn
\section{Appendix}

\subsection{Dataset information}
\label{Dataset_information}

MIMIC-IV (version 3.1) contains longitudinal hospital
and ICU records from Beth Israel Deaconess Medical
Center. Each hospital admission or ICU stay constitutes
a separate prediction instance, meaning that a patient
may contribute multiple instances if they have multiple
admissions or stays.

We evaluate four binary prediction tasks. For
\textbf{ICU mortality}, prediction is made at ICU
admission, with the outcome defined as death during
the corresponding ICU stay. For \textbf{ICU
readmission}, prediction is made at ICU discharge,
with the outcome indicating a subsequent ICU
admission. Similarly, \textbf{hospital mortality}
is predicted at hospital admission, with the outcome
defined as death during that hospitalisation, while
\textbf{30-day hospital readmission} is predicted
at hospital discharge, with the outcome indicating
readmission within 30 days.

For each task, only clinical events available up to
the corresponding prediction time are included.
Train, validation, and test partitions are constructed
at the patient level using unique subject IDs, ensuring
that all admissions and stays belonging to the same
patient remain within a single partition.

\FloatBarrier
\begin{table}[H]
\centering
\caption{MIMIC-IV Population and Task-Specific Distribution}
\label{tab:mimic_combined}
\small
\begin{tabular}{lcccccc}
\toprule
& \multicolumn{2}{c}{\textbf{Train}} & \multicolumn{2}{c}{\textbf{Validation}} & \multicolumn{2}{c}{\textbf{Test}} \\
\cmidrule(lr){2-3} \cmidrule(lr){4-5} \cmidrule(lr){6-7}
\textbf{Metric / Task} & \textbf{Count} & \textbf{\%} & \textbf{Count} & \textbf{\%} & \textbf{Count} & \textbf{\%} \\
\midrule
\textit{Unique Patients} & 291,702 & 80.0\% & 36,463 & 10.0\% & 36,462 & 10.0\% \\
\midrule
\textbf{ICU Mortality} & & & & & & \\
\quad Case (Positive) & 5,881 & 7.8\% & 705 & 7.7\% & 715 & 7.7\% \\
\quad Control (Negative) & 69,169 & 92.2\% & 8,508 & 92.3\% & 8,628 & 92.3\% \\
\midrule
\textbf{ICU Readmission} & & & & & & \\
\quad Case (Positive) & 10,745 & 14.2\% & 1,345 & 14.5\% & 1,350 & 14.3\% \\
\quad Control (Negative) & 64,978 & 85.8\% & 7,947 & 85.5\% & 8,079 & 85.7\% \\
\midrule
\textbf{Hospital Mortality} & & & & & & \\
\quad Case (Positive) & 8,061 & 2.9\% & 1,014 & 2.9\% & 988 & 2.8\% \\
\quad Control (Negative) & 271,056 & 97.1\% & 34,271 & 97.1\% & 33,840 & 97.2\% \\
\midrule
\textbf{Hospital Readmission} & & & & & & \\
\quad Case (Positive) & 86,859 & 20.4\% & 11,074 & 20.6\% & 11,480 & 21.2\% \\
\quad Control (Negative) & 339,559 & 79.6\% & 42,659 & 79.4\% & 42,676 & 78.8\% \\
\bottomrule
\end{tabular}
\end{table}

\FloatBarrier
\subsection{Baselines and Ablations}
\label{Baselins_and_Ablations}

\paragraph{XGBoost}
We implement XGBoost using bag-of-codes frequency features constructed from all discrete medical concepts observed in each patient's available history. Each feature corresponds to a clinical code and records the number of times that code occurs before the task-specific prediction time. This provides a non-sequential structured-data baseline that retains event frequency but discards temporal ordering, continuous values, and relationships between events. 

\paragraph{Transformer Encoder}
We implement a code-based Transformer encoder, similar to \citet{Placido2023ATrajectories}, with 12 encoder layers, 8 attention heads, and a maximum context length of 2,048 events. The model operates directly on symbolic medical codes rather than natural-language descriptions, with each code becoming a unique token. Temporal information and continuous measurements are discretised into additional learned representations following the code-based preprocessing pipeline.

The encoder follows an analogous two-stage training procedure to tEHR. In Stage~1, it is pretrained on trajectories from the training split using masked language modelling with a masking probability of 15\%; of selected tokens, 80\% are replaced by the mask token, 10\% by a random token, and 10\% are left unchanged, following BERT \citep{Devlin2019BERT}. In Stage~2, a linear classification head is added to the patient representation and the encoder is fine-tuned jointly for each downstream prediction task using binary cross-entropy. We train both stages for 50 epochs with early stopping on validation loss.

\paragraph{CLMBR-T}
We evaluate CLMBR-T \citep{Wornow2023EHRSHOT:Models}, a 141M-parameter structured-EHR Transformer pretrained on 2.57 million patients, on the four MIMIC-IV tasks. Because CLMBR-T operates over a predefined structured vocabulary, MIMIC-IV concepts were mapped using the public MIMIC-IV-to-OMOP pipeline \citep{OHDSIMIMICConversion}. This mapped 79.4\% of observed event occurrences, with unmapped events omitted from the input sequence. We evaluate two variants. In the frozen setting, CLMBR-T is used as a fixed patient representation and only a logistic-regression classification head is trained. In the fine-tuned setting, the pretrained model is additionally adapted to each downstream task before classification. Because the mapping is incomplete, these results should be interpreted as a transfer benchmark under partial vocabulary alignment rather than as a comparison under fully matched input representations.

\paragraph{BioClinical-ModernBERT} We use BioClinical-ModernBERT-base (150M parameters) \cite{Sounack2025BioClinicalNLP} as a pretrained medical text encoder and fine-tune all model parameters for each downstream prediction task. BioClinical-ModernBERT receives the same textualised EHR representation as tEHR. Its maximum context length is 8,192 tokens; trajectories exceeding this limit are truncated from the left to retain the most recent clinical history, while demographic information is preserved at the beginning of the sequence. A linear binary classification head is applied to the patient-level representation and all parameters are optimised jointly with the encoder using binary cross-entropy.

\paragraph{Zero-shot LLM}
To evaluate whether pretrained LLM knowledge alone is sufficient for clinical prediction, we use the instruction-tuned version of the corresponding LLM without any parameter updates. The model is provided with the textualised patient trajectory together with a task-specific instruction requesting a binary prediction, encoded as \texttt{0} for control and \texttt{1} for case. Predictions are obtained from the model's output logits for the two candidate labels, which are normalised to produce a continuous case score for AUROC and AUPRC evaluation. No examples or demonstrations are included in the prompt. The patient trajectory is truncated only when it exceeds the context window of the underlying LLM.

\paragraph{Prompt-based Supervised Fine-tuning}
The prompt-based SFT baseline uses the same textualised
input representation and binary prediction prompt as
the zero-shot setting, but adapts the LLM to the
labelled downstream cohort using LoRA. Unlike tEHR,
no continued pretraining on unlabelled EHR trajectories
is performed, and no discriminative classification
head is introduced. Instead, the model is trained to
generate the correct binary outcome token from the
patient trajectory. This evaluates generative
supervised fine-tuning as an alternative to tEHR's
discriminative adaptation of the patient-level
representation.

The prompts for both are as follows:

\begin{Verbatim}[breaklines=true, breakanywhere=true, fontsize=\footnotesize]
You are an expert in clinical prediction from longitudinal electronic health record (EHR) data.
Your task is to predict whether the patient will {task} based only on the EHR information provided below.
Demographic information appears at the beginning, followed by clinical events in chronological order. Events are separated by semicolons. Numerical measurements are reported to one decimal place with their units, and elapsed time between events is included.
Consider the patient's longitudinal history, including relevant clinical events, measurements, and trends over time.
Answer with exactly one digit on a single line:
1 = yes
0 = no
Do not output any additional text, explanation, probability, or symbols.

Patient's medical history:
{ehr_text}

Answer:
\end{Verbatim}

\paragraph{Mistral-7B with Atomic Clinical-Code Tokens} 
To assess whether tEHR's performance can be explained
by model capacity rather than its use of natural-language
event descriptions, we construct a capacity-controlled
variant using the same Mistral-7B backbone. Each unique
clinical concept is represented by an atomic token,
rather than its natural-language description. We extend
the Mistral vocabulary with the clinical-code vocabulary
used by the code-based Transformer baseline, initialising
new embeddings for these codes.

The model follows the same two-stage training procedure
as tEHR. During Stage~1, the LoRA adapters and newly
introduced code embeddings are updated through
autoregressive next-token continued pretraining on
patient trajectories. During Stage~2, a linear
classification head is introduced, and the LoRA
adapters, code embeddings, and classification head
are optimised jointly using binary cross-entropy.
This preserves the backbone and adaptation procedure
while replacing natural-language descriptions with
learned symbolic code representations.

\paragraph{LLM-initialised Transformer} 
Motivated by the use of pretrained language-model
embeddings for medical concepts in
\citet{Park2025EnhancingEmbeddings}, we evaluate whether pretrained language semantics
can be incorporated into a conventional code-based
EHR Transformer through its input embedding layer.
This variant uses the same architecture and
structured event representation as the Transformer
Encoder baseline, but initialises clinical-code
embeddings using pretrained LLM representations
of their corresponding natural-language descriptions.
Each description is independently encoded by the
LLM to obtain a semantic representation used to
initialise the corresponding code embedding, which
remains trainable during subsequent adaptation.

Following initialisation, the model uses the same
two-stage training procedure as the Transformer
Encoder baseline: masked language modelling on
unlabelled patient trajectories followed by
task-specific supervised fine-tuning using binary
cross-entropy. The remaining architecture and
training procedure are unchanged, allowing
pretrained semantic information to be introduced
without processing full natural-language trajectories
through the longitudinal model.

\paragraph{Event LLM Embeddings}
To evaluate the utility of pretrained language representations without adapting the LLM itself, we encode each textualised clinical event independently using the frozen base LLM to obtain event-level embeddings. These embeddings are then passed to a trainable six-layer, eight-head Transformer encoder that models the longitudinal patient sequence and produces a patient-level representation for binary classification. The LLM parameters remain fixed throughout, while the Transformer aggregator and classification head are trained on the labelled cohort. This provides an event-level representation baseline analogous to \cite{Hur2024UniHPFKnowledge}.

\paragraph{LLM Projection} 
We evaluate an alternative approach in which
pretrained event-level representations are supplied
directly to the longitudinal LLM rather than
processing clinical descriptions as natural-language
token sequences. Each clinical event is independently
encoded using a frozen pretrained LLM to obtain an
event-level embedding. These embeddings are projected
into the input embedding space of Mistral-7B,
allowing each clinical event to occupy a single
position in the longitudinal sequence rather than
multiple subword tokens.

The projected event sequence is processed by
Mistral-7B to produce a patient-level representation
for binary classification. Unlike tEHR, this variant
does not undergo autoregressive next-token continued
pretraining on longitudinal EHR trajectories.
Instead, it is adapted directly to each downstream
prediction task using a supervised classification
objective. This separates event-level semantic
encoding from longitudinal aggregation while
reducing the sequence expansion associated with
natural-language textualisation.

\subsection{Hyperparameter search}
The same patient-level train/validation/test splits are used as for tEHR and the different hyperparameters for each of the models can be seen in Table \ref{tab:hyperparams}.

\begin{table}[h]
\centering
\caption{Hyperparameter Search Space for Baseline Models.}
\label{tab:hyperparams}
\small 
\begin{tabular}{lll}
\toprule
\textbf{Model} & \textbf{Hyperparameter} & \textbf{Search Space / Values} \\
\midrule
\textbf{XGBoost} & Learning Rate ($\eta$) & $\{0.01, \mathbf{0.05}, 0.1, 0.2\}$ \\
& Max Depth & $\{5, \mathbf{6}, 7, 8, 9, 10\}$ \\
& Subsample Ratio & $\{0.6, 0.8, \mathbf{1.0}\}$ \\
& Gamma & log-uniform [\text{0.1}, 5.0] \\
\midrule
\textbf{BioClinical-ModernBERT} & Learning Rate & $\{\mathbf{1e-5}, 2e-5, 5e-5\}$ \\
& Weight Decay & $\{\mathbf{0.01}, 0.1\}$ \\
& Dropout Rate & $\{\mathbf{0.1}, 0.2, 0.3\}$ \\
& Batch Size & $\{16, \mathbf{32}\}$ \\
\midrule
\textbf{Transformer Encoder} & Hidden Dimension ($d_{model}$) & $\{256, 512, \mathbf{768}, 1024\}$ \\
& Number of Layers & $\{4, 6, 8, \mathbf{12}\}$ \\
& Attention Heads & $\{4, \mathbf{8}\}$ \\
\midrule
\textbf{Main experiments} & LoRA Rank ($r$) & $\{8, \mathbf{16}, 32\}$ \\
(All LoRA models) & LoRA Alpha ($\alpha$) & $\{16, \mathbf{32}, 64\}$ \\
& Learning Rate & $\{5e-5, \mathbf{1e-4}, 2e-4\}$ \\
& Weight Decay & $\{\mathbf{0.01}, 0.1\}$ \\
& Dropout & $\{\mathbf{0.1}, 0.2\}$ \\
\bottomrule
\end{tabular}
\end{table}

\FloatBarrier
\subsection{Calibration}
\label{app:calibration}

We additionally assess the calibration of predicted risks
on the held-out MIMIC-IV test sets. Table~\ref{tab:calibration}
reports Brier scores and calibration slopes for tEHR and
the strongest competing baselines. tEHR achieves the
lowest Brier score across all four tasks, although this
metric reflects both discrimination and calibration.
Its calibration slopes are closest to the ideal value
of 1 on three of the four tasks, with particularly
good calibration for ICU and hospital readmission.

\begin{table}[h]
\centering
\caption{
Brier scores and calibration slopes across four MIMIC-IV
prediction tasks. Lower Brier scores are better, while
calibration slopes closer to 1 indicate better calibration
slope. Best values are shown in bold.
}
\label{tab:calibration}

\begin{tabular}{@{}lccccc@{}}
\toprule
\textbf{Task}
& \shortstack{\textbf{tEHR}\\\textbf{Mistral-7B}}
& \shortstack{\textbf{Transformer}\\\textbf{Encoder}}
& \shortstack{\textbf{CLMBR-T}\\\textbf{fine-tuned}}
& \shortstack{\textbf{LLM-initialised}\\\textbf{Transformer}}
& \textbf{XGBoost} \\
\midrule

\multicolumn{6}{@{}l}{\textit{Brier score}} \\
ICU mortality
& \textbf{0.055} & 0.067 & 0.063 & 0.060 & 0.061 \\
ICU readmission
& \textbf{0.091} & 0.102 & 0.107 & 0.096 & 0.106 \\
Hospital mortality
& \textbf{0.022} & 0.025 & 0.024 & 0.023 & 0.024 \\
Hospital readmission
& \textbf{0.131} & 0.146 & 0.181 & 0.137 & 0.142 \\

\midrule
\multicolumn{6}{@{}l}{\textit{Calibration slope}} \\
ICU mortality
& \textbf{0.90} & 0.66 & 0.49 & 0.70 & 1.12 \\
ICU readmission
& \textbf{1.01} & 1.25 & 0.57 & 0.70 & 1.32 \\
Hospital mortality
& 0.85 & 1.07 & 0.66 & \textbf{1.01} & 1.06 \\
Hospital readmission
& \textbf{1.03} & 0.78 & 0.74 & 0.84 & 1.16 \\

\bottomrule
\end{tabular}
\end{table}

\FloatBarrier
\subsection{Tokenisation and context length}
\label{context_length}

Textualising structured EHR events expands each clinical
event into multiple language-model subword tokens.
On MIMIC-IV training trajectories, we measured a median
expansion of 7.41 Mistral tokens per clinical event
(Table~\ref{tab:bpe_per_event}).
To prevent differences in available clinical history
from confounding model comparisons, we apply a shared
event-level truncation procedure. For each patient,
we retain the most recent clinical events, up to a
maximum of 2,048 events, subject to the corresponding
textualised sequence fitting within tEHR's
16,000-token context window. The same retained
clinical events are then supplied to the code-based
and semantic representation models using their
respective input representations. Demographic
information is preserved at the beginning of each
trajectory.
The exception is BioClinical-ModernBERT, whose
8,192-token context window may require additional
truncation of the matched textualised trajectory, and CLMBR-T, whose 496 token context window limits the history of events.
For this model, demographic information and the
most recent events are retained within its
context limit.


\begin{table}[H]
\centering
\caption{Mistral BPE tokens per clinical event on MIMIC-IV training
trajectories, measured after textualising the most recent 2,048 events
with temporal information and numerical values retained. Percentiles are
computed within each task and averaged equally across the four tasks.}
\label{tab:bpe_per_event}
\begin{tabular}{lccc}
\toprule
 & 25th pct. & Median & 75th pct. \\
\midrule
BPE / event & 7.04 & 7.41 & 7.81 \\
\bottomrule
\end{tabular}
\end{table}

\FloatBarrier
\subsection{MIMIC-IV Full results}


\begin{table}[H]
\centering

\begingroup
\newcommand{\metric}[2]{%
    \makecell[c]{$#1_{\pm #2}$}%
}
\newcommand{\metricbf}[2]{%
    \makecell[c]{$\mathbf{#1}_{\pm #2}$}%
}

\caption{
MIMIC-IV baseline, ablation, and tEHR results.
Point estimates are test-set AUROC / AUPRC presented as mean $\pm$
standard deviation across 3 seeds.
Ablations use tEHR-Mistral-7B-v0.3 unless otherwise specified.
}
\label{tab:mimic_results_sd}

\renewcommand{\arraystretch}{1.15}

\resizebox{\textwidth}{!}{%
\begin{tabular}{lcccccccc}
\toprule
& \multicolumn{2}{c}{ICU Mortality}
& \multicolumn{2}{c}{ICU Readmission}
& \multicolumn{2}{c}{Hospital Mortality}
& \multicolumn{2}{c}{Hospital Readmission} \\
\cmidrule(lr){2-3}
\cmidrule(lr){4-5}
\cmidrule(lr){6-7}
\cmidrule(lr){8-9}

\textbf{Model / experiment}
& \textbf{AUROC} & \textbf{AUPRC}
& \textbf{AUROC} & \textbf{AUPRC}
& \textbf{AUROC} & \textbf{AUPRC}
& \textbf{AUROC} & \textbf{AUPRC} \\
\midrule

\multicolumn{9}{l}{\textbf{Baselines}} \\

XGBoost
& \metric{0.817}{0.0009} & \metric{0.306}{0.0053}
& \metric{0.713}{0.0008} & \metric{0.333}{0.0018}
& \metric{0.868}{0.0007} & \metric{0.235}{0.0031}
& \metric{0.742}{0.0010} & \metric{0.483}{0.0028} \\

Transformer Encoder
& \metric{0.866}{0.0010} & \metric{0.370}{0.0056}
& \metric{0.773}{0.0009} & \metric{0.456}{0.0016}
& \metric{0.911}{0.0007} & \metric{0.281}{0.0040}
& \metric{0.765}{0.0013} & \metric{0.531}{0.0025} \\

CLMBR-T (frozen)
& \metric{0.793}{0.0004} & \metric{0.235}{0.0031}
& \metric{0.724}{0.0006} & \metric{0.340}{0.0009}
& \metric{0.881}{0.0004} & \metric{0.204}{0.0022}
& \metric{0.708}{0.0005} & \metric{0.409}{0.0015} \\

CLMBR-T (fine-tuned)
& \metric{0.861}{0.0005} & \metric{0.406}{0.0038}
& \metric{0.762}{0.0009} & \metric{0.428}{0.0011}
& \metric{0.915}{0.0006} & \metric{0.328}{0.0029}
& \metric{0.762}{0.0007} & \metric{0.515}{0.0022} \\

BioClinical-ModernBERT
& \metric{0.824}{0.0011} & \metric{0.300}{0.0052}
& \metric{0.739}{0.0010} & \metric{0.359}{0.0021}
& \metric{0.860}{0.0008} & \metric{0.197}{0.0041}
& \metric{0.766}{0.0012} & \metric{0.532}{0.0030} \\

Zero-shot LLM
& \metric{0.559}{0.0019} & \metric{0.109}{0.0074}
& \metric{0.539}{0.0019} & \metric{0.147}{0.0035}
& \metric{0.627}{0.0015} & \metric{0.048}{0.0050}
& \metric{0.528}{0.0020} & \metric{0.232}{0.0046} \\

Prompt SFT -- Mistral-7B-v0.3
& \metric{0.811}{0.0017} & \metric{0.221}{0.0063}
& \metric{0.668}{0.0018} & \metric{0.219}{0.0032}
& \metric{0.786}{0.0012} & \metric{0.073}{0.0044}
& \metric{0.724}{0.0016} & \metric{0.437}{0.0043} \\

\midrule
\multicolumn{9}{l}{\textbf{Ablations: alternative semantic injection}} \\

LLM-initialised Transformer
& \metric{0.843}{0.0007} & \metric{0.351}{0.0049}
& \metric{0.778}{0.0008} & \metric{0.467}{0.0020}
& \metric{0.897}{0.0008} & \metric{0.275}{0.0039}
& \metric{0.766}{0.0011} & \metric{0.532}{0.0028} \\

Event LLM embeddings
& \metric{0.739}{0.0008} & \metric{0.194}{0.0060}
& \metric{0.547}{0.0014} & \metric{0.166}{0.0028}
& \metric{0.876}{0.0009} & \metric{0.215}{0.0049}
& \metric{0.739}{0.0012} & \metric{0.480}{0.0029} \\

LLM projection
& \metric{0.737}{0.0009} & \metric{0.210}{0.0055}
& \metric{0.651}{0.0009} & \metric{0.273}{0.0026}
& \metric{0.817}{0.0010} & \metric{0.156}{0.0036}
& \metric{0.737}{0.0009} & \metric{0.483}{0.0033} \\

\midrule
\multicolumn{9}{l}{\textbf{Ablations: training framework}} \\

No Stage~1
& \metric{0.843}{0.0014} & \metric{0.347}{0.0016}
& \metric{0.727}{0.0107} & \metric{0.326}{0.0020}
& \metric{0.900}{0.0009} & \metric{0.265}{0.0014}
& \metric{0.768}{0.0005} & \metric{0.522}{0.0009} \\

Frozen stage 1 model
& \metric{0.794}{0.0015} & \metric{0.240}{0.0018}
& \metric{0.735}{0.0018} & \metric{0.366}{0.0021}
& \metric{0.871}{0.0010} & \metric{0.207}{0.0015}
& \metric{0.726}{0.0005} & \metric{0.462}{0.0011} \\

Frozen base LLM
& \metric{0.778}{0.0016} & \metric{0.222}{0.0019}
& \metric{0.718}{0.0019} & \metric{0.335}{0.0023}
& \metric{0.848}{0.0011} & \metric{0.189}{0.0016}
& \metric{0.703}{0.0005} & \metric{0.418}{0.0010} \\

Mistral-7B atomic codes
& \metric{0.772}{0.0008} & \metric{0.221}{0.0055}
& \metric{0.703}{0.0009} & \metric{0.323}{0.0019}
& \metric{0.871}{0.0006} & \metric{0.226}{0.0032}
& \metric{0.766}{0.0010} & \metric{0.511}{0.0023} \\

\midrule
\multicolumn{9}{l}{\textbf{Ablations: clinical information representation}} \\

Hashed concepts
& \metric{0.766}{0.0016} & \metric{0.245}{0.0020}
& \metric{0.702}{0.0018} & \metric{0.299}{0.0022}
& \metric{0.873}{0.0009} & \metric{0.218}{0.0015}
& \metric{0.736}{0.0005} & \metric{0.463}{0.0010} \\

Scrambled concepts
& \metric{0.791}{0.0015} & \metric{0.285}{0.0019}
& \metric{0.705}{0.0018} & \metric{0.295}{0.0022}
& \metric{0.869}{0.0010} & \metric{0.226}{0.0015}
& \metric{0.728}{0.0005} & \metric{0.450}{0.0010} \\

No elapsed-time text
& \metric{0.887}{0.0012} & \metric{0.428}{0.0015}
& \metric{0.794}{0.0016} & \metric{0.499}{0.0018}
& \metric{0.931}{0.0008} & \metric{0.346}{0.0012}
& \metric{0.772}{0.0005} & \metric{0.540}{0.0009} \\

No numerical values
& \metric{0.851}{0.0013} & \metric{0.346}{0.0017}
& \metric{0.759}{0.0017} & \metric{0.411}{0.0021}
& \metric{0.913}{0.0009} & \metric{0.297}{0.0014}
& \metric{0.752}{0.0005} & \metric{0.481}{0.0010} \\

\midrule
\multicolumn{9}{l}{\textbf{tEHR}} \\

tEHR-Qwen3-0.6B
& \metric{0.822}{0.0010} & \metric{0.289}{0.0052}
& \metric{0.738}{0.0011} & \metric{0.414}{0.0023}
& \metric{0.903}{0.0009} & \metric{0.273}{0.0041}
& \metric{0.759}{0.0010} & \metric{0.522}{0.0037} \\

tEHR-Qwen3-8B
& \metric{0.870}{0.0011} & \metric{0.391}{0.0044}
& \metric{0.780}{0.0009} & \metric{0.480}{0.0016}
& \metric{0.927}{0.0015} & \metric{0.324}{0.0039}
& \metric{0.764}{0.0012} & \metric{0.527}{0.0025} \\

tEHR-Mistral-7B-v0.3
& \metricbf{0.887}{0.0008} & \metricbf{0.424}{0.0048}
& \metricbf{0.794}{0.0008} & \metricbf{0.498}{0.0015}
& \metricbf{0.932}{0.0007} & \metricbf{0.333}{0.0035}
& \metricbf{0.775}{0.0009} & \metricbf{0.556}{0.0024} \\

\bottomrule
\end{tabular}%
}

\endgroup
\end{table}

\FloatBarrier
\subsection{Additional Metrics}
We report precision, recall, F1, balanced accuracy, accuracy, and Matthews correlation coefficient (MCC) for all MIMIC-IV tasks, as indicated in the table below. Because these metrics require a decision threshold, the threshold for each model and task was selected exclusively on the validation set by maximising F1 score, and was then fixed before evaluation on the test set. This avoids tuning the operating point on the test data. Across the four tasks, tEHR-Mistral-7B achieves the highest F1 and MCC in all four settings, while other models occasionally attain higher recall or accuracy depending on the selected operating point.

\begin{table*}[h]
\centering
\caption{ICU mortality. Thresholds maximise validation F1 and are then fixed and applied to the test set. Best test-set value for each metric is shown in bold.}
\label{tab:operating_icu_mortality}
\begin{tabular}{lccccccc}
\toprule
\textbf{Model} & \textbf{Thr.} & \textbf{Prec.} & \textbf{Rec.} & \textbf{F1} & \textbf{Bal.\ acc.} & \textbf{Acc.} & \textbf{MCC} \\
\midrule
XGBoost & 0.180 & 0.331 & 0.399 & 0.362 & 0.666 & 0.892 & 0.305 \\
BioClinical-ModernBERT & 0.222 & 0.323 & 0.411 & 0.362 & 0.670 & 0.889 & 0.304 \\
Transformer Encoder & 0.156 & 0.363 & 0.484 & 0.415 & 0.707 & 0.895 & 0.363 \\
Zero-shot LLM & 0.150 & 0.168 & 0.179 & 0.174 & 0.553 & 0.870 & 0.103 \\
Prompt SFT & 0.400 & 0.223 & \textbf{0.660} & 0.333 & 0.735 & 0.798 & 0.298 \\
CLMBR-T (frozen) & 0.722 & 0.273 & 0.392 & 0.322 & 0.653 & 0.874 & 0.259 \\
CLMBR-T (fine-tuned) & 0.097 & 0.337 & 0.580 & 0.426 & \textbf{0.743} & 0.880 & 0.382 \\
tEHR-Mistral-7B & 0.245 & \textbf{0.422} & 0.485 & \textbf{0.451} & 0.715 & \textbf{0.910} & \textbf{0.403} \\
\bottomrule
\end{tabular}
\end{table*}

\begin{table*}[h]
\centering
\caption{ICU readmission. Thresholds maximise validation F1 and are then fixed and applied to the test set. Best test-set value for each metric is shown in bold.}
\label{tab:operating_icu_readmission}
\begin{tabular}{lccccccc}
\toprule
\textbf{Model} & \textbf{Thr.} & \textbf{Prec.} & \textbf{Rec.} & \textbf{F1} & \textbf{Bal.\ acc.} & \textbf{Acc.} & \textbf{MCC} \\
\midrule
XGBoost & 0.165 & 0.285 & 0.435 & 0.344 & 0.633 & 0.777 & 0.224 \\
BioClinical-ModernBERT & 0.154 & 0.305 & 0.521 & 0.385 & 0.668 & 0.776 & 0.273 \\
Transformer Encoder & 0.149 & 0.403 & 0.483 & 0.440 & 0.686 & 0.835 & 0.346 \\
Zero-shot LLM & 0.240 & 0.151 & 0.727 & 0.250 & 0.546 & 0.414 & 0.066 \\
Prompt SFT & 0.480 & 0.222 & 0.548 & 0.316 & 0.625 & 0.681 & 0.181 \\
CLMBR-T (frozen) & 0.594 & 0.319 & 0.453 & 0.375 & 0.652 & 0.797 & 0.263 \\
CLMBR-T (fine-tuned) & 0.284 & 0.381 & 0.454 & 0.414 & 0.670 & 0.827 & 0.316 \\
tEHR-Mistral-7B & 0.217 & \textbf{0.430} & 0.514 & \textbf{0.468} & \textbf{0.704} & \textbf{0.843} & \textbf{0.379} \\
\bottomrule
\end{tabular}
\end{table*}

\begin{table*}[h]
\centering
\caption{Hospital mortality. Thresholds maximise validation F1 and are then fixed and applied to the test set. Best test-set value for each metric is shown in bold.}
\label{tab:operating_hospital_mortality}
\begin{tabular}{lccccccc}
\toprule
\textbf{Model} & \textbf{Thr.} & \textbf{Prec.} & \textbf{Rec.} & \textbf{F1} & \textbf{Bal.\ acc.} & \textbf{Acc.} & \textbf{MCC} \\
\midrule
XGBoost & 0.185 & 0.340 & 0.292 & 0.314 & 0.638 & \textbf{0.964} & 0.297 \\
BioClinical-ModernBERT & 0.093 & 0.228 & 0.385 & 0.286 & 0.673 & 0.946 & 0.270 \\
Transformer Encoder & 0.134 & 0.331 & 0.351 & 0.341 & 0.665 & 0.962 & 0.321 \\
Zero-shot LLM & 0.230 & 0.049 & 0.381 & 0.087 & 0.584 & 0.776 & 0.068 \\
Prompt SFT & 0.234 & 0.088 & \textbf{0.555} & 0.151 & 0.694 & 0.825 & 0.167 \\
CLMBR-T (frozen) & 0.858 & 0.237 & 0.387 & 0.294 & 0.675 & 0.948 & 0.277 \\
CLMBR-T (fine-tuned) & 0.068 & 0.338 & 0.444 & 0.384 & 0.709 & 0.960 & 0.367 \\
tEHR-Mistral-7B & 0.202 & \textbf{0.363} & 0.447 & \textbf{0.401} & \textbf{0.712} & 0.962 & \textbf{0.384} \\
\bottomrule
\end{tabular}
\end{table*}

\FloatBarrier
\begin{table*}[h]
\centering
\caption{Hospital readmission. Thresholds maximise validation F1 and are then fixed and applied to the test set. Best test-set value for each metric is shown in bold.}
\label{tab:operating_hospital_readmission}
\begin{tabular}{lccccccc}
\toprule
\textbf{Model} & \textbf{Thr.} & \textbf{Prec.} & \textbf{Rec.} & \textbf{F1} & \textbf{Bal.\ acc.} & \textbf{Acc.} & \textbf{MCC} \\
\midrule
XGBoost & 0.203 & 0.370 & 0.638 & 0.468 & 0.673 & 0.693 & 0.293 \\
BioClinical-ModernBERT & 0.232 & 0.418 & 0.607 & 0.495 & 0.690 & 0.737 & 0.336 \\
Transformer Encoder & 0.155 & 0.403 & 0.638 & 0.494 & 0.692 & 0.723 & 0.332 \\
Zero-shot LLM & 0.240 & 0.235 & \textbf{0.772} & 0.361 & 0.549 & 0.420 & 0.087 \\
Prompt SFT & 0.321 & 0.326 & 0.736 & 0.452 & 0.663 & 0.621 & 0.267 \\
CLMBR-T (frozen) & 0.540 & 0.361 & 0.583 & 0.446 & 0.653 & 0.693 & 0.264 \\
CLMBR-T (fine-tuned) & 0.521 & 0.416 & 0.613 & 0.496 & 0.691 & 0.736 & 0.337 \\
tEHR-Mistral-7B & 0.266 & \textbf{0.457} & 0.594 & \textbf{0.517} & \textbf{0.702} & \textbf{0.764} & \textbf{0.370} \\
\bottomrule
\end{tabular}
\end{table*}


%% file: references.bib
@article{Placido2023ATrajectories,
    title = {{A deep learning algorithm to predict risk of pancreatic cancer from disease trajectories}},
    year = {2023},
    journal = {Nature Medicine},
    author = {Placido, Davide and Yuan, Bo and Hjaltelin, Jessica X. and Zheng, Chunlei and Haue, Amalie D. and Chmura, Piotr J. and Yuan, Chen and Kim, Jihye and Umeton, Renato and Antell, Gregory and Chowdhury, Alexander and Franz, Alexandra and Brais, Lauren and Andrews, Elizabeth and Marks, Debora S. and Regev, Aviv and Ayandeh, Siamack and Brophy, Mary T. and Do, Nhan V. and Kraft, Peter and Wolpin, Brian M. and Rosenthal, Michael H. and Fillmore, Nathanael R. and Brunak, Søren and Sander, Chris},
    number = {5},
    month = {5},
    pages = {1113--1122},
    volume = {29},
    doi = {10.1038/s41591-023-02332-5},
    issn = {1078-8956}
}

@article{Makarov2025ARecords,
    title = {{A transformer-based model for next disease prediction using electronic health records}},
    year = {2025},
    journal = {The European Physical Journal Special Topics},
    author = {Makarov, Nikolai and Lipkovich, Mikhail},
    month = {1},
    doi = {10.1140/epjs/s11734-024-01447-1},
    issn = {1951-6355}
}

@article{Xu2025ACES:Datasets,
    title = {{ACES: Automatic Cohort Extraction System for Event-Stream Datasets}},
    year = {2025},
    author = {Xu, Justin and Gallifant, Jack and Johnson, Alistair E. W. and McDermott, Matthew B. A.},
    month = {3},
    arxivId = {2406.19653}
}

@inproceedings{Devlin2019BERT,
    title = {{BERT}},
    year = {2019},
    booktitle = {Proceedings of the 2019 Conference of the North},
    author = {Devlin, Jacob and Chang, Ming-Wei and Lee, Kenton and Toutanova, Kristina},
    pages = {4171--4186},
    publisher = {Association for Computational Linguistics},
    address = {Stroudsburg, PA, USA},
    doi = {10.18653/v1/N19-1423}
}

@article{Sounack2025BioClinicalNLP,
    title = {{BioClinical ModernBERT: A State-of-the-Art Long-Context Encoder for Biomedical and Clinical NLP}},
    year = {2025},
    author = {Sounack, Thomas and Davis, Joshua and Durieux, Brigitte and Chaffin, Antoine and Pollard, Tom J. and Lehman, Eric and Johnson, Alistair E. W. and McDermott, Matthew and Naumann, Tristan and Lindvall, Charlotta},
    month = {6},
    arxivId = {2506.10896}
}

@article{Shoham2024CPLLM:Models,
    title = {{CPLLM: Clinical Prediction with Large Language Models}},
    year = {2024},
    author = {Shoham, Ofir Ben and Rappoport, Nadav},
    month = {5},
    arxivId = {2309.11295}
}

@article{Ding2026DiscreteModeling,
    title = {{Discrete Neural Interlingua for Ontology-Agnostic EHR Modeling}},
    year = {2026},
    author = {Ding, Tong and Tian, Caiwei and Bang, Dongmin and Lu, Ming Y. and Wagner, Sophia J. and Zhang, Andrew and Pettit, Rowland and Le, Long Phi and Mahmood, Faisal},
    month = {10},
    arxivId = {2610.27047}
}

@article{Fallahpour2024EHRMamba:Records,
    title = {{EHRMamba: Towards Generalizable and Scalable Foundation Models for Electronic Health Records}},
    year = {2024},
    author = {Fallahpour, Adibvafa and Alinoori, Mahshid and Ye, Wenqian and Cao, Xu and Afkanpour, Arash and Krishnan, Amrit},
    month = {11},
    arxivId = {2405.14567}
}

@article{Wornow2023EHRSHOT:Models,
    title = {{EHRSHOT: An EHR Benchmark for Few-Shot Evaluation of Foundation Models}},
    year = {2023},
    author = {Wornow, Michael and Thapa, Rahul and Steinberg, Ethan and Fries, Jason A. and Shah, Nigam H.},
    month = {12},
    arxivId = {2307.02028}
}

@article{Park2025EnhancingEmbeddings,
    title = {{Enhancing EHR-based pancreatic cancer prediction with LLM-derived embeddings}},
    year = {2025},
    journal = {npj Digital Medicine},
    author = {Park, Jiheum and Patterson, Jason and Acitores Cortina, Jose M. and Gu, Tian and Hur, Chin and Tatonetti, Nicholas},
    number = {1},
    month = {7},
    pages = {465},
    volume = {8},
    doi = {10.1038/s41746-025-01869-8},
    issn = {2398-6352}
}

@article{Chandak2026EveryQuery:Records,
    title = {{EveryQuery: Zero-Shot Clinical Prediction via Task-Conditioned Pretraining over Electronic Health Records}},
    year = {2026},
    author = {Chandak, Payal and Kondas, Gregory and Friedman, Liat Antwarg and Kohane, Isaac and McDermott, Matthew},
    month = {5},
    arxivId = {2603.07900}
}

@article{Kraljevic2023ForesightEHRs,
    title = {{Foresight -- Generative Pretrained Transformer (GPT) for Modelling of Patient Timelines using EHRs}},
    year = {2023},
    author = {Kraljevic, Zeljko and Bean, Dan and Shek, Anthony and Bendayan, Rebecca and Hemingway, Harry and Yeung, Joshua Au and Deng, Alexander and Baston, Alfie and Ross, Jack and Idowu, Esther and Teo, James T and Dobson, Richard J},
    month = {1},
    arxivId = {2212.08072}
}

@article{Rajamohan2025FoundationScale,
    title = {{Foundation Models for Clinical Records at Health System Scale}},
    year = {2025},
    author = {Rajamohan, Haresh Rengaraj and Gao, Xiang and Zhu, Weicheng and Huang, Shih-Lun and Chen, Long and Cho, Kyunghyun and Deniz, Cem M. and Razavian, Narges},
    month = {7},
    arxivId = {2507.00574}
}

@article{Waxler2025GenerativeScale,
    title = {{Generative Medical Event Models Improve with Scale}},
    year = {2025},
    author = {Waxler, Shane and Blazek, Paul and White, Davis and Sneider, Daniel and Chung, Kevin and Nagarathnam, Mani and Williams, Patrick and Voeller, Hank and Wong, Karen and Swanhorst, Matthew and Zhang, Sheng and Usuyama, Naoto and Wong, Cliff and Naumann, Tristan and Poon, Hoifung and Loza, Andrew and Meeker, Daniella and Hain, Seth and Shah, Rahul},
    month = {11},
    arxivId = {2508.12104}
}

@misc{WorldHealthOrganizationInternationalICD,
    title = {{International Statistical Classification of Diseases and Related Health Problems (ICD)}},
    author = {{World Health Organization}},
    howpublished = {https://www.who.int/standards/classifications/classification-of-diseases}
}

@article{Hegselmann2026LargeEncoders,
    title = {{Large language models are powerful electronic health record encoders}},
    year = {2026},
    journal = {npj Digital Medicine},
    author = {Hegselmann, Stefan and von Arnim, Georg and Rheude, Tillmann and Kronenberg, Noel and Sontag, David and Hindricks, Gerhard and Eils, Roland and Wild, Benjamin},
    number = {1},
    month = {7},
    pages = {530},
    volume = {9},
    doi = {10.1038/s41746-026-02915-9},
    issn = {2398-6352}
}

@article{Shmatko2025LearningTransformers,
    title = {{Learning the natural history of human disease with generative transformers}},
    year = {2025},
    journal = {Nature},
    author = {Shmatko, Artem and Jung, Alexander Wolfgang and Gaurav, Kumar and Brunak, Søren and Mortensen, Laust Hvas and Birney, Ewan and Fitzgerald, Tom and Gerstung, Moritz},
    month = {9},
    doi = {10.1038/s41586-025-09529-3},
    issn = {0028-0836}
}

@article{Schulman2025LoRARegret,
    title = {{LoRA Without Regret}},
    year = {2025},
    journal = {Thinking Machines Lab: Connectionism},
    author = {Schulman, John and Lab, Thinking Machines},
    doi = {10.64434/tml.20250929}
}

@article{Hu2021LoRA:Models,
    title = {{LoRA: Low-Rank Adaptation of Large Language Models}},
    year = {2021},
    author = {Hu, Edward J. and Shen, Yelong and Wallis, Phillip and Allen-Zhu, Zeyuan and Li, Yuanzhi and Wang, Shean and Wang, Lu and Chen, Weizhu},
    month = {10},
    arxivId = {2106.09685}
}

@article{BertArnrich2024MedicalHealth,
    title = {{Medical Event Data Standard ({\{}MEDS{\}}): Facilitating Machine Learning for Health}},
    year = {2024},
    journal = {ICLR 2024 Workshop on Learning from Time Series For Health},
    author = {{Bert Arnrich} and {Edward Choi} and {Jason Alan Fries} and {Matthew B.A. McDermott} and {Jungwoo Oh} and {Tom Pollard} and {Nigam Shah} and {Ethan Steinberg} and {Michael Wornow} and {Robin van de Water}}
}

@misc{OHDSIMIMICConversion,
    title = {{MIMIC IV to OMOP CDM Conversion}},
    author = {{OHDSI}},
    howpublished = {https://github.com/OHDSI/MIMIC}
}

@article{Johnson2024MIMIC-IV,
    title = {{MIMIC-IV}},
    year = {2024},
    journal = {PhysioNet},
    author = {Johnson, Alistair and Bulgarelli, Lucas and Pollard, Tom and Gow, Brian and Moody, Benjamin and Horng, Steven and Celi, Leo Anthony and Mark, Roger},
    month = {10},
    url = {https://doi.org/10.13026/kpb9-mt58},
    doi = {10.13026/kpb9-mt58}
}

@misc{McDermott2025MIMIC-IVFormat,
    title = {{MIMIC-IV MEDS: An ETL pipeline to extract MIMIC-IV data into the MEDS format}},
    year = {2025},
    author = {McDermott, Matthew and Xu, Justin},
    month = {4},
    publisher = {Zenodo},
    url = {https://github.com/Medical-Event-Data-Standard/MIMIC_IV_MEDS},
    doi = {10.5281/zenodo.17535580}
}

@article{Steinberg2023MOTOR:Records,
    title = {{MOTOR: A Time-To-Event Foundation Model For Structured Medical Records}},
    year = {2023},
    author = {Steinberg, Ethan and Fries, Jason and Xu, Yizhe and Shah, Nigam},
    month = {12},
    arxivId = {2301.03150}
}

@article{Harutyunyan2019MultitaskData,
    title = {{Multitask learning and benchmarking with clinical time series data}},
    year = {2019},
    author = {Harutyunyan, Hrayr and Khachatrian, Hrant and Kale, David C. and Steeg, Greg Ver and Galstyan, Aram},
    month = {8},
    doi = {10.1038/s41597-019-0103-9},
    arxivId = {1703.07771}
}

@misc{NHSEnglandPrimaryCareDomain2025Non-TypeDMNONTYPE1_COD,
    title = {{Non-Type 1 Diabetes Mellitus (DMNONTYPE1{\_}COD)}},
    year = {2025},
    author = {{NHS England Primary Care Domain}},
    publisher = {OpenCodelists},
    url = {https://www.opencodelists.org/codelist/nhsd-primary-care-domain-refsets/dmnontype1_cod/20250912/}
}

@article{Zhu2024PromptingData,
    title = {{Prompting Large Language Models for Zero-Shot Clinical Prediction with Structured Longitudinal Electronic Health Record Data}},
    year = {2024},
    author = {Zhu, Yinghao and Wang, Zixiang and Gao, Junyi and Tong, Yuning and An, Jingkun and Liao, Weibin and Harrison, Ewen M. and Ma, Liantao and Pan, Chengwei},
    month = {2},
    arxivId = {2402.01713}
}

@article{Lee2026RepresentationModels,
    title = {{Representation Before Training: A Fixed-Budget Benchmark for Generative Medical Event Models}},
    year = {2026},
    author = {Lee, Inhyeok and Solo, Luke and Burkhart, Michael C. and Ramadan, Bashar and Parker, William F. and Beaulieu-Jones, Brett K.},
    month = {4},
    arxivId = {2604.16775}
}

@article{Attrach2025RethinkingMore,
    title = {{Rethinking Tokenization for Clinical Time Series: When Less is More}},
    year = {2025},
    author = {Attrach, Rafi Al and Fani, Rajna and Restrepo, David and Jia, Yugang and Sch{\"{u}}ffler, Peter},
    month = {12},
    arxivId = {2512.05217}
}

@article{Munoz-Farre2022SEHR-CE:Expansion,
    title = {{sEHR-CE: Language modelling of structured EHR data for efficient and generalizable patient cohort expansion}},
    year = {2022},
    author = {Munoz-Farre, Anna and Rose, Harry and Cakiroglu, Sera Aylin},
    month = {11},
    arxivId = {2211.17121}
}

@article{Stearns2001SNOMEDStatus.,
    title = {{SNOMED clinical terms: overview of the development process and project status.}},
    year = {2001},
    journal = {Proceedings. AMIA Symposium},
    author = {Stearns, M Q and Price, C and Spackman, K A and Wang, A Y},
    pages = {662--6},
    issn = {1531-605X},
    pmid = {11825268}
}

@article{Chisholm1990TheClassification.,
    title = {{The Read clinical classification.}},
    year = {1990},
    journal = {BMJ},
    author = {Chisholm, J},
    number = {6732},
    month = {4},
    pages = {1092--1092},
    volume = {300},
    doi = {10.1136/bmj.300.6732.1092},
    issn = {0959-8138}
}

@article{Guo2026TokenizationModels,
    title = {{Tokenization Tradeoffs in Structured EHR Foundation Models}},
    year = {2026},
    author = {Guo, Lin Lawrence and Arciniegas, Santiago Eduardo and Lee, Joseph Jihyung and Yan, Adam Paul and Tomlinson, George and Fries, Jason and Sung, Lillian},
    month = {3},
    arxivId = {2603.15644}
}

@article{Wang2024Transformer-basedData,
    title = {{Transformer-based deep learning model for the diagnosis of suspected lung cancer in primary care based on electronic health record data}},
    year = {2024},
    journal = {eBioMedicine},
    author = {Wang, Lan and Yin, Yonghua and Glampson, Ben and Peach, Robert and Barahona, Mauricio and Delaney, Brendan C. and Mayer, Erik K.},
    month = {12},
    pages = {105442},
    volume = {110},
    doi = {10.1016/j.ebiom.2024.105442},
    issn = {23523964}
}

@article{Hur2024UniHPFKnowledge,
    title = {{UniHPF : Universal Healthcare Predictive Framework with Zero Domain Knowledge}},
    year = {2024},
    author = {Hur, Kyunghoon and Oh, Jungwoo and Kim, Junu and Kim, Jiyoun and Lee, Min Jae and Cho, Eunbyeol and Moon, Seong-Eun and Kim, Young-Hak and Choi, Edward},
    month = {9},
    arxivId = {2211.08082}
}

@misc{DanielHan2023Unsloth,
    title = {{Unsloth}},
    year = {2023},
    author = {Daniel Han, Michael Han and team, Unsloth},
    url = {https://github.com/unslothai/unsloth}
}

@article{Renc2024ZeroTransformer,
    title = {{Zero shot health trajectory prediction using transformer}},
    year = {2024},
    journal = {npj Digital Medicine},
    author = {Renc, Pawel and Jia, Yugang and Samir, Anthony E. and Was, Jaroslaw and Li, Quanzheng and Bates, David W. and Sitek, Arkadiusz},
    number = {1},
    month = {9},
    pages = {256},
    volume = {7},
    doi = {10.1038/s41746-024-01235-0},
    issn = {2398-6352}
}
